\documentclass{article}

\usepackage{arxiv}

\usepackage[utf8]{inputenc} % allow utf-8 input
\usepackage[T1]{fontenc}    % use 8-bit T1 fonts
\usepackage{hyperref}       % hyperlinks
\usepackage{url}            % simple URL typesetting
\usepackage{booktabs}       % professional-quality tables
\usepackage{amsfonts}       % blackboard math symbols
\usepackage{nicefrac}       % compact symbols for 1/2, etc.
\usepackage{microtype}      % microtypography
\usepackage{lipsum}
\usepackage{graphicx}
\usepackage{subfiles}
\usepackage{comment}

\usepackage{biblatex} %Imports biblatex package
\graphicspath{ {./images/} }

\title{Latin writing styles analysis with Machine Learning\\ New approach to old questions}

\author{

 Arianna Di Bernardo  \\
  Department of Physics, University of Turin\\
  Machine Learning Journal Club, Turin\\
  \texttt{arianna.dibernardo@edu.unito.it} \\
   \And
 Simone Poetto  \\
  Department of Physics, University of Turin\\
  Machine Learning Journal Club, Turin\\
  \texttt{simone.poetto@edu.unito.it} \\
  \And
 Sillano Pietro \\
  Department of Physics, University of Turin\\
  Machine Learning Journal Club, Turin\\
  \texttt{pietro.sillano@edu.unito.it} \\
 \And
 Beatrice Villata \\
  Department of Physics, University of Turin\\
  Machine Learning Journal Club, Turin\\
  \texttt{beatrice.villata@edu.unito.it} \\
 \And
 Weronika Sójka  \\
  Faculty of Philosophy and Social Sciences \\
  Nicolaus Copernicus University of Torun \\
  Neurotechtor Scientific Students Club \\
  \texttt{w.sojka00@gmail.com}
  \And
 Zofia Piętka-Danilewicz  \\
  Faculty of Philosophy and Social Sciences \\
  Nicolaus Copernicus University of Torun \\
  Neurotechtor Scientific Students Club \\
  \texttt{pietkadanilewicz@gmail.com}
  \AND
 Piotr Pranke  \\
  Department of History \\
  Nicolaus Copernicus University of Torun \\
  \texttt{piotrpranke@umk.pl} \\
}

\begin{document}
\maketitle
\begin{abstract}
In the Middle Ages texts were learned by heart and spread using oral means of communication from generation to generation. Adaptation of the art of prose and poems allowed keeping particular descriptions and compositions characteristic for many literary genres. Taking into account such a specific construction of literature composed in Latin, we can search for and indicate the probability patterns of familiar sources of specific narrative texts. Consideration of Natural Language Processing tools allowed us the transformation of textual objects into numerical ones and then application of machine learning algorithms to extract information from the dataset. We carried out the task consisting of the practical use of those concepts and observation to create a tool for analyzing narrative texts basing on open-source databases. The tool focused on creating specific search tools resources which could enable us detailed searching throughout the text. The main objectives of the study take into account finding similarities between sentences and between documents. Next, we applied machine learning algorithms on chosen texts to calculate specific features of them (for instance authorship or centuries) and to recognize sources of anonymous texts with a certain percentage.
%\end{abstract}

% keywords can be removed

\end{abstract}

\vspace{20pt}

All code is avaiable here: \url{https://github.com/MachineLearningJournalClub/THUS-Torun}

\newpage
\section{Introduction}

Writing texts in the Middle Ages required sticking to specific rules of composition. The variety of medieval rhetoric providing instruction in the production of letters and documents is referred to as \textit{ars dictandi}, deriving from the Latin verb \textit{dictare} - "to compose". The term was used during the Middle Ages to designate any composition types, and it was based on the assumption that authors had to produce their texts in line with well-worked models \cite{emil1992ars}.\\

Detecting the sources of a Latin text and inferring their historical ages or the authors whose writing styles are comparable would be fundamental for the reasons stated above. Anonymous documents could be matched with the most likely author to obtain more specific basic information about the context. That would allow an understanding of a more extensive historical and cultural context as give a deeper insight into international relations.\\ 
 
A way to achieve this goal is employing tools from Natural Language Processing (NLP), a Machine Learning (ML) field that aims to analyze text and extract useful information from them \cite{zhang2016semantic}. NLP has proved to be the best way for Topic Extraction, Sentiment Analysis, and Text Summarization in which we are the most interested in \cite{li2012literature}, \cite{Pang2002ThumbsUS}.\\

However, this proves to be a challenge for our purpose, as most of the existing NLP tools are developed for modern languages. Historical language often differs significantly from its modern counterpart in several aspects that make a naive application of these tools problematic  \cite{bollmann2018normalization} \cite{piotrowski2012natural}.\\

Some steps have been done in the direction of building NLP tools to preprocess Latin texts \cite{cltk}, for words lemmatization and part-of-speech tagging with Deep Learning tools \cite{kestemont2016integrated}, for text categorization with Latin etymologies \cite{fang2009latin}, as well as for identifying intertextual relationships between authors \cite{burns2021profiling}. However, none of these technologies has yet been used to classify the origin of entire Latin texts.\\ 

There are ongoing efforts to digitize historical documents to help their preservation and to improve their accessibility to further the possibility of discovering interesting features \cite{piotrowski2012natural}. The emerging field of \textit{digital humanities} aims to exploit these opportunities by combining traditional methods with computer-based tools as information retrieval, lemmatizers, part-of-speech taggers, data mining, visualization, and geographic information systems \cite{berti2019digital}. However, there has been surprisingly little communication between fields of humanities computing and computational methods. Due to the need to create an effortful application for historical documents analysis, we suggest producing a different way to investigate given data. This work aims to generate a model that can perform the recognition of authors in the Latin language. On this basis, we extracted three annotated samples of Latin documents from the \textit{Latin Library} database (\url{https://www.thelatinlibrary.com/}). Suggested work may be a further step in the field of digital humanities.\\
\section{Research Questions}
The study's objective is to investigate the employment of a general pipeline to extract qualitative and quantitative information from Latin texts operating ML algorithms. In particular, our approach focuses on detecting stylistic features of the texts. A semantic NLP approach is proposed to find general characteristics. That may help with capturing domain knowledge, for illustration, by finding similarities between documents. That could be helpful for historians and linguists to investigate the technical features of a text without processing all of them itself. It could be, for example, beneficial to classify authors according to writing styles or historical ages. Equally, it would be valuable to hypothesize the authorship of anonymous documents basing similarities with authors of the same period or location. \\ 

Using different methods, we aim to explore the structure of texts and the differences between authors. The main questions we want to answer are:

\begin{itemize}
    \item May ML methods help find a way to classify a group of texts by authors?
    \item Is it possible to match anonymous documents to the authors with similar styles?
\end{itemize}

Additionally, a network representation that aims to capture and constitute the relationship between different Latin authors has been proposed. 
\section{Methods}

We developed our analysis in different steps: first, by considering NLP tools to transform textual objects into numerical ones, an operation called embedding. Then by applying some well-known machine learning algorithms to extract useful information from the dataset.

\subsection{Natural Language Processing}
Natural Language Processing, or NLP, is a computational technique that allows language analysis. During this work, NLP has been used to analyze Latin texts. In particular, we handled some examinations that are usually carried out manually in ancient languages.\\
The most common tools to carry out NLP analysis work on modern languages only. For this reason, analyzing ancient texts with machine learning is usually a demanding job.
Additionally, another challenge encountered is the unavailability of ancient works in text formats. Usually, OCR techniques are needed, and it is not easy to find sources online. However, Latin Library provides a database of text files. In this way, it has not been necessary to focus on the translation step. \\

\begin{figure}[h] 
\centering 
\includegraphics[scale=0.6]{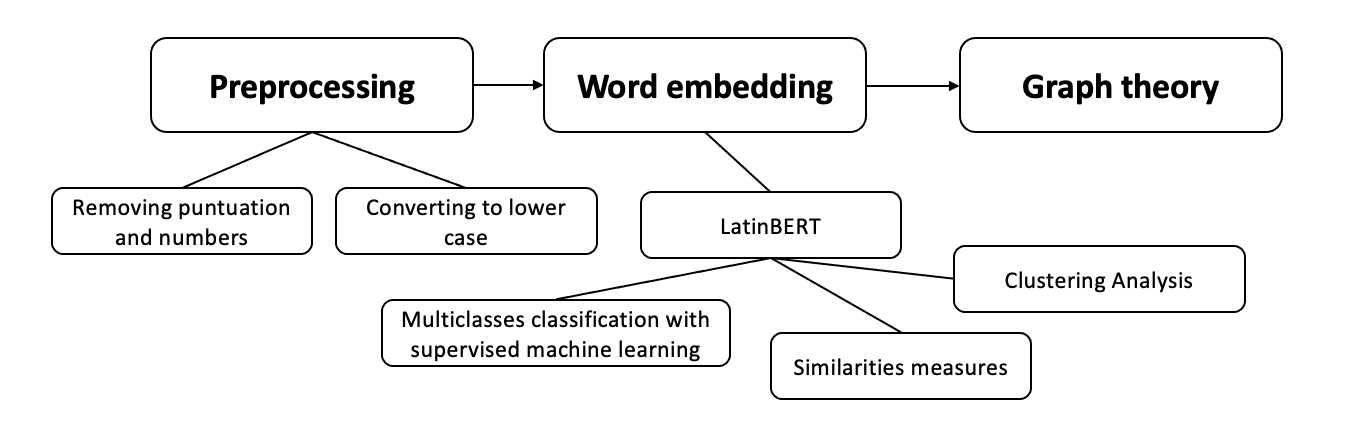} 
\caption{Information extraction methodology followed.}
\label{img:graph} 
\end{figure}

LatinBERT allowed us to improve our analysis since it was a pre-trained model to work on ancient Latin. \cite{bollmann2018normalization}\\
In Figure \ref{img:graph}, the steps followed to extract information from the text are summarized.

\subsubsection{Preprocessing}

The \textit{preprocessing} phase is conducted to prepare the text for further analysis. The goal is to clean the raw text and make it consistent, to apply algorithms for the embedding.
Cleaning means, for instance, handling punctuation, removing uninformative words or correcting spelling errors. \\
%To achieve this result, we employed \textit{cltk}, a python library that is specifically developed to deal with classical languages. \\
In particular, the preprocessing operation applied in this analysis are the following:
\begin{itemize}
\item Removing numbers and uninformative words: numbers do not possess semantic meaning, so they are removed. The same goes for misspelt words that the algorithm is not able to correct referring to a standard Latin vocabulary.
\item Converting to lower cases to standardize the text.
\item Removing punctuation: for analyzing whole documents we also remove the punctuation, while we retain it when we analyzed the texts at the sentence level.
\end{itemize}

\subsubsection{Word embedding}

The computational analysis of written documents requires the transformation of the text into mathematical objects on which we can perform operations. This is done by assigning a unique numerical vector to each word or sentence, creating a projection of the terms in a high dimensional vector space.
We refer to this process as \textit{word embedding} \cite{turian-etal-2010-word}. \\

Numerous algorithms in the literature can be employed to achieve this goal \cite{w2v}, \cite{d2v}, \cite{bert}, but little has been done about ancient languages. In the present work, we make use of \textbf{LatinBERT}, a contextual language model for Latin, trained on 642.7 million words from various sources from the classical era to the 21st century (\textit{Corpus Thomisticum, Internet Archive, Latin Library, Patrologia Latina, Perseus, Latin Wikipedia}) \cite{bamman2020latin}. \\

LatinBERT, as all BERT models, produces contextual embeddings of the words. Hence, each term is expressed as a unique vector, and its value depends on the context in which the word appears. BERT models take all the words appearing in a sentence and infer the relationship between them. This approach allows the model to deal with polysemic expressions, associating them with diverse embeddings when representing different concepts.\\
BERT models have proven to be more effective than other embedding techniques because, when they compute the embedding for a word, they take into consideration all previous and next words in the same sentence.
The model was made available by the authors for future research in the field.\\ 

\subsection{Similarities measures}

By applying Latin-Bert it is possible to encode documents into vectors and then exploit their geometrical properties to define similarity between the native texts. The most intuitive way to extract information about the closeness of two vectors in space is through the \textit{Euclidean distance}, which naturally represents the shortest distance between two points.\\

An alternative way to quantify such closeness is by computing the \textit{cosine similarity} $\phi$ of pairs of vectors, a metric often used in NLP to measure similarity between words or texts. It is measured by the cosine of the angle $\theta$ between two vectors $V$ and $W$, and it determines whether the two vectors point in the same direction \cite{datam}:

\begin{equation}
\phi =\cos (\theta)=\frac{\mathbf{v} \cdot \mathbf{w}}{\|\mathbf{v}\|\|\mathbf{w}\|}=\frac{\sum_{i=1}^{n} v_{i} w_{i}}{\sqrt{\sum_{i=1}^{n} v_{i}^{2}} \sqrt{\sum_{i=1}^{n} w_{i}^{2}}}.
\end{equation}

Here $v_{i}$ and $w_{i}$ are the components of the $n-$dimensional vectors $\textbf{v}$ and $\textbf{w}$, in our case standing for the word embedding vectors of the documents.\\

The value of $\phi$ ranges between -1 and 1. Values of $\phi$ closer to -1 indicate dissimilarity, while values of $\phi$ closer to 1 mean similarity between the two objects. If $\phi$ is close to zero, then no significant correlation between the two texts exists.\\

\subsection{Clustering Analysis}

Complementary with computing similarity measures between texts, we can group them based on their closeness in the multi-dimensional space they lie with \textit{clustering analysis}. Clustering algorithms belong to the general class of Unsupervised ML techniques and apply when the objects divide into natural groups, referred to as clusters. There exist numerous algorithms that can be used for this purpose. The one we selected is one of the most popular, the \textit{K-means} algorithm. It consists of grouping the vectors by evaluating the euclidean distance between them, resulting in the assignment of each data point to the nearest cluster \cite{10.5555/1162264}.

\subsection{Visualization}

A further step that must be taken into account when working with digital humanities is making the results intuitive and intelligible. The best way to deal with such interpretability is to create a graphical representation of the results.\\

A way to visualize documents in such a high-dimensional numerical form is employing \textit{dimensionality reduction}. Since the visualization of points in dimensions 4 and above is impossible, it's necessary to reduce them to be represented as points in dimensions 2 or 3.

In our work, we applied the \textit{Uniform Manifold Approximation and Projection} (UMAP) embedding algorithm to reduce to two the dimension of the embedding vectors. UMAP exploits geometrical and topological features of the vectors to preserves their global structure: the embedding is found by searching for a low dimensional projection of the data that has the closest possible equivalent topological structure \cite{mcinnes2018umap}. Once we obtain such low-dimensional projection, we can visualize the vectors in the plane and observe how they cluster and their relationships.\\

The best way to represent relationships between objects is by employing graphs. A graph $G$ is an ordered pair $(V, E)$, where $V$ is a set of nodes and $E$ is a set of edges. Nodes represent items, while edges represent relationships between them and can be weighted with geometric measures capturing the degree of such relationships, such as the cosine similarity or the euclidean distance previously introduced. The information about the connectivity patterns and the weights are collected in the \textit{connectivity matrix} $W$ of the graph.
In the present study, nodes represent authors, while the links represent the connection between them and are weighted with their degree of similarity. 
\section{Results}

In the present work, three sets of Latin text fragments from different authors were analyzed. Latin text samples were taken from authors from diverse centuries and works. \\

The first part of the analysis was the preprocessing part. That is useful to normalize the text by removing punctuation and numbers and converting words to the lower cases. \\

An example of this technique is observable by the sentence \textit{"Populus autem eodem anno me consulem, cum cos uterque bello cecidisset, et triumvirum rei publicae constituendae creavit."} that is preprocessed as \textit{"populus autem eodem anno me consulem cum cos uterque bello cecidisset et triumvirum rei publicae constituendae creavit"}.\\

Preprocessed sentences have been then modified with Latin-BERT in their numerical counterpart. The primary analysis has been performed in two different ways: firstly, converting the sentences into numerical objects, secondly processing the entire document. 

That results in two sets of 768-dimensional vectors collected in two matrices.

Additionally, information about the author of the texts in two \textit{label vectors} has been collected where each element is represented by an integer number and standing for the specific author. \\

Once the vectors were collected, we carried on the following analysis.

\subsection{Group texts by authors}

After the pre-processing, as stated previously, we obtain two different sets of vectors: one for the conversion of single tenses and another for entire documents.\\
It is then possible to extract similarities between sentences and documents through clustering analysis on the embedded texts. The expectation was that the numerical counterpart of sentences or documents from the same author clusterized across the vectorial space.\\

The K-means algorithm creates a fit, computing the labels for each vector.  That is performed on the two sets of vectors separately. The algorithm performance is evaluated by checking the correspondence between the K-means labels and the author labels, collected in the label vectors previously defined. We obtain an accuracy of 0.91 on the first set and an accuracy of 0.96 on the second. The above results clearly show that both entire documents or just short sentences clusterize concerning the author who wrote them.\\

\begin{figure}[h]
    \centering 
    \includegraphics[scale=0.6]{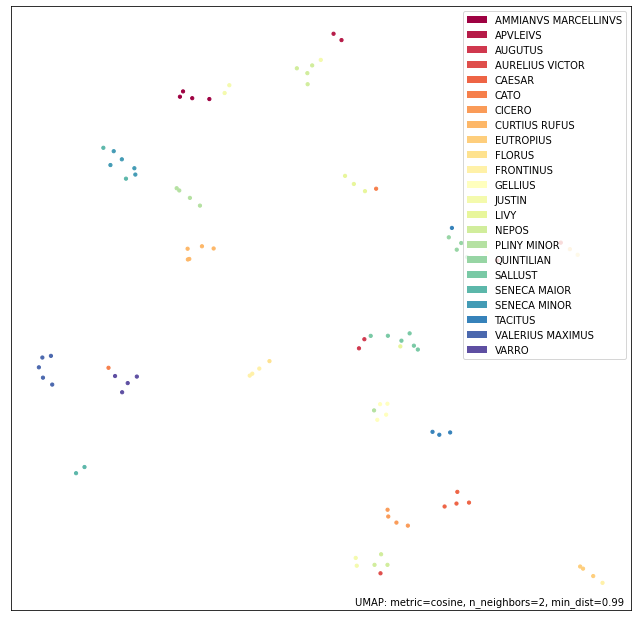}
    \caption{Clusters of documents by authors: Each point represent a text and each color correspond to an author. }
    \label{fig1}
\end{figure}

\begin{figure}
    \centering 
    \includegraphics[scale=1.1]{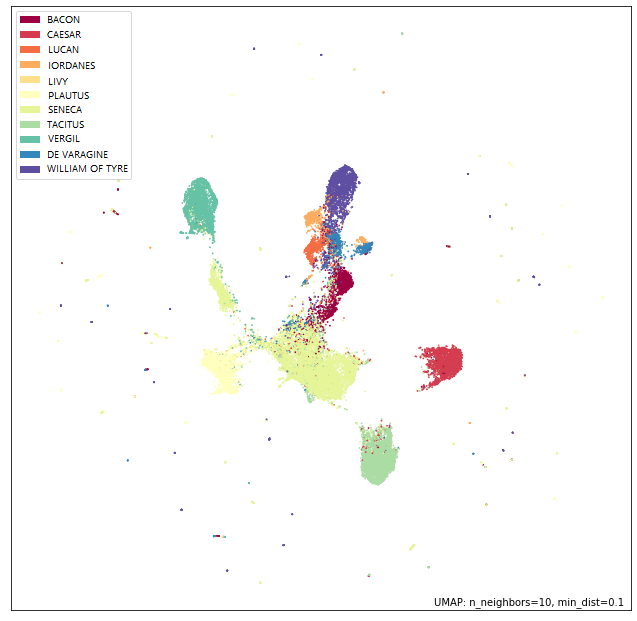}
    \caption{Clusters of sentences by authors: Each point represent a sentence and each color correspond to an author. }
    \label{fig2}
\end{figure}

To show such a tendency to cluster based on authorship, we plot the embedded vectors in a lower-dimensional space using UMAP, as shown in Figure \ref{fig1} and \ref{fig2}.

\subsection{Authors' relationships network}

The relationships between authors have been captured by creating a network representation of the similarities. The vectors usually cluster according to the authorship, as it has been previously verified on the single sentences and the entire documents cases.\\
For this analysis, the focus has been placed on the second sample of vectors. In particular, the set of the 23 K-means centroids has been involved. Those centroids depict the central points of each cluster and can be considered as vectors representing the author classes.\\

After computing the cosine similarity for each pair of centroids and collecting the values in a 23$\times$23 matrix the matrix is normalized with the Min-Max Scaling to make relationships more evident. \\

\begin{figure} 
    \centering 
    \includegraphics[scale=0.5]{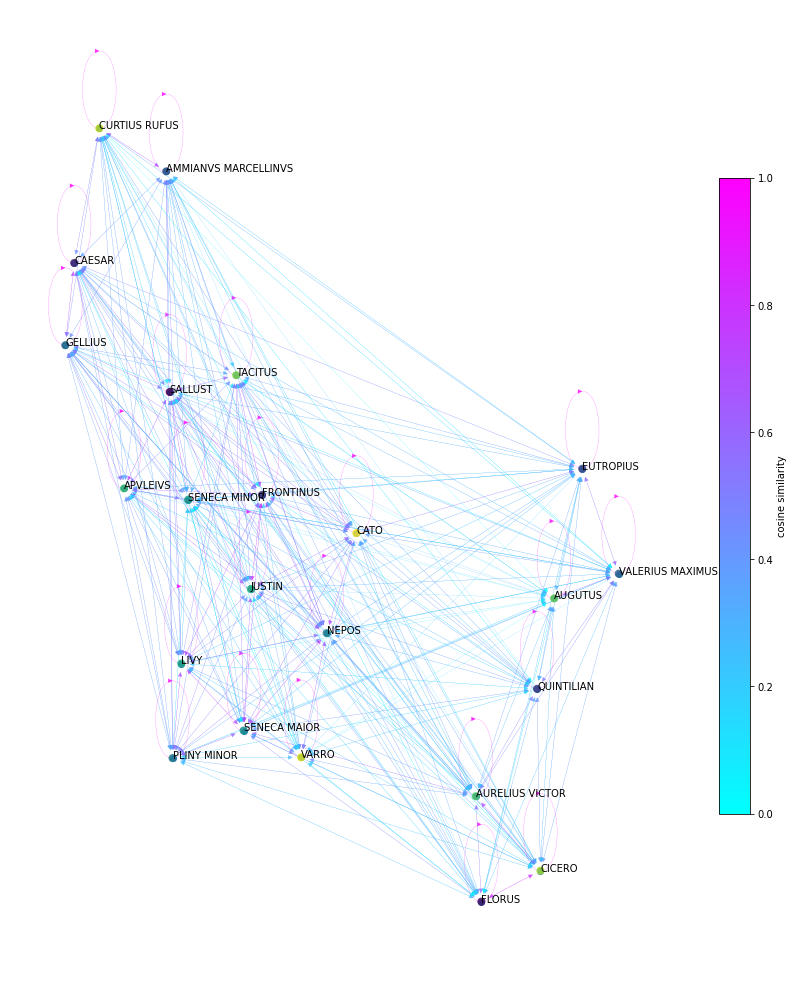}
    \caption{Network representation of the relationship between authors. Link are weighted with the cosine similarity measure.}
    \label{fig3}
\end{figure}

\begin{figure} 
    \centering 
    \includegraphics[scale=0.5]{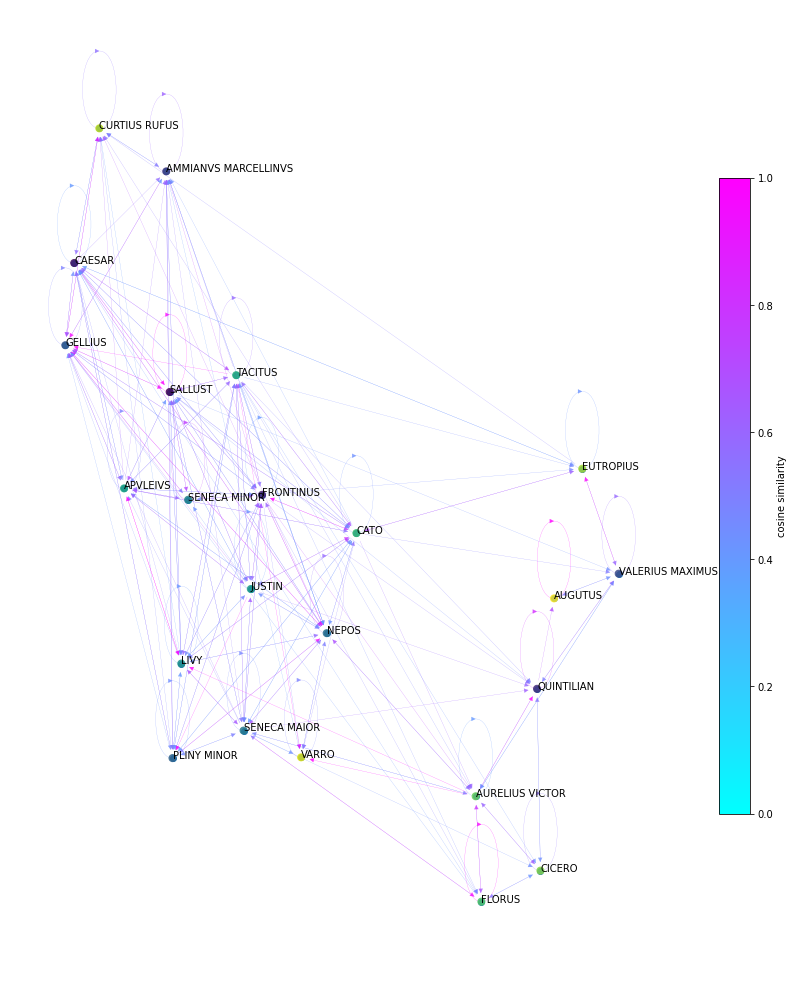}
    \caption{Network representation of the relationship between authors. Only stronger connections are shown.}
    \label{fig4}
\end{figure}

Finally, the rescaling cosine similarity matrix is used as a weighted connectivity matrix to create the network capturing the relationship between authors (\ref{fig3}). In a second stage, a threshold at 0.7 allows displaying only the most relevant connections between authors (\ref{fig4}).

\subsection{Author recognition for Gesta Principum Polonorum}

In the present section, we deal with the question of authorship of the \textit{Gesta Principum Polonorum}. The unknown author is referred to as 'Gallus Anonymus' and there are different hypotheses about who is.

Original Gallus Anonimus texts do not exist anymore since they have been lost at an unknown time (supposedly around the 12th century, because then Gallus repeats the errors proper to the later manuscripts) \cite{budkowa1953anonima}. Originally there were 8 Chronicles, while now only their authorized copies are available. However, the most significant information for linguists is that they were not multiplied mechanically, since they date back before the popularization of printing. Replication by hand changed the overall structure because it adds personal perspective according to the time and individuals that perform this operation. 

Many medieval works argue about the nationality and origin of Gallus Anonymous. Cognitive values resulting from the determination of the author's origin would allow them to add a historical message to his texts. One of the hypotheses is based on similarities observed between the Gesta and the text from an author known as the ‘Monk of Lido’ from Venice. 
The text considered is the 'Translatio Sancti Nicolai' \cite{eder2015search}. \\

\begin{figure} 
    \centering 
    \includegraphics[scale=0.5]{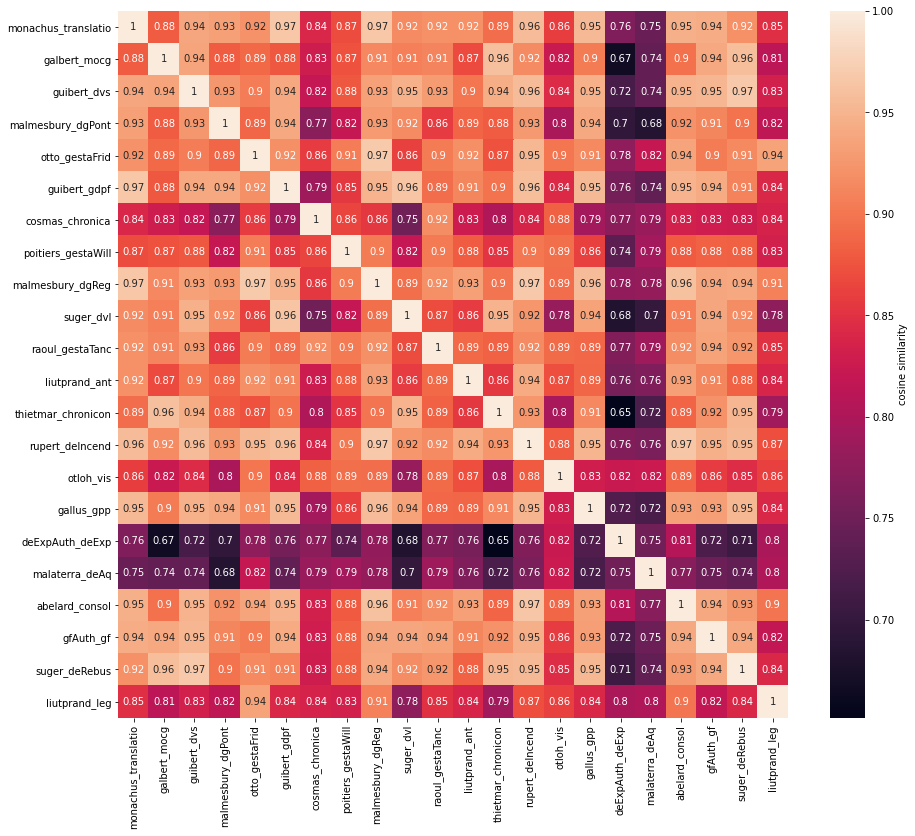}
    \caption{Cosine similarity matrix.}
    \label{fig5}
\end{figure}

To confirm these hypotheses and find new similarities, the following analysis has been developed. Taking a set of 39 full texts of varying lengths (7897–173,536 words) and genres (history, theology, political theory) written by 22 different authors (counting Gallus and Monk as separate authors), they are transformed into single sentences from these texts in their vectorial form by using Latin-BERT, as previously mentioned.
That produces 2200 vectors of $768$ components, afterwards fitted the K-means algorithm on them. Since the accuracy of the algorithm is still high (0.91), it is possible to compute the centroids for each cluster considering them as representative elements of the author classes. The cosine similarity between each pair of K-means centroids was collected in a $22 \times 22 $ matrix (\ref{fig5}).\\

The above computation indicates a 0.95 similarity between Gallus Anonymus and Monk of Lido. It also points out a high similarity between Gallus Anonymus and William of Malmesbury, Suger, and Rupert.

The cited results not only confirm the hypothesis of the attribution of the Gesta Principum Polonorum to the Monk of Lido but also brings up other authorship attribution possibilities which could be taken into account in further historical studies.

\section{Conclusions}
During the work exposed, different methods have been applied to Latin texts to conduct analysis and find similarities using NLP and machine learning techniques.\\
In particular, LatinBERT has been involved instead of word2vec or doc2vec, used in previous works \cite{burns2021profiling}. \\
By using a pre-trained model, it has been possible to conduct a different kind of analysis without having to find sources to train the model.\\

The final product of this work is a pipeline that could be followed to extract information from ancient texts. In particular, this paper focuses on Latin texts, but the same could be easily applied to other languages by changing the preprocessing phase. In fact, after this step, the sentences are embedded into mathematical objects and placed into a geometrical space to detect closeness between these vectors. This closeness, which is detected by clustering the dataset and applying the K-means method, gives information about the similarities of the objects in the text.\\

The first step conducted consist of a method to detect the authorship of a text. That is carried out by detecting similarities and clustering with a metric that considers similarities between authors. %By applying k-means it is possible to extract an accuracy using a supervised method. 
In this way, we conclude that our model can cluster the text we involved with high accuracy ($ >90\%$). Additionally, with a similarity matrix, it is also possible to gain information about similarities between authors.\\

Another method developed to underline the connection between authors is the relationship network, a network representation that displays the bonds between the centroids of the clusters.\\ 

Finally, we applied the analysis we performed to shed light with a computational approach on the open question of the authorship of the anonymous \textit{Gesta Principum Polonorum}. Our results strongly support the previous hypothesis of attribution of the document to the Monk of Lido.
%, and open new paths of authorship to be evaluated from a historical point of view.

\printbibliography %Prints bibliography

@misc{emil1992ars,
  title={Ars dictaminis, ars dictandi, Typologie des sources du Moyen Age occidental, Fasc. 60},
  author={Emil. J. Polak},
  year={1992},
  publisher={JSTOR}
}

@article{piotrowski2012natural,
  title={Natural language processing for historical texts},
  author={Piotrowski, Michael},
  journal={Synthesis lectures on human language technologies},
  volume={5},
  number={2},
  pages={1--157},
  year={2012},
  publisher={Morgan \& Claypool Publishers}
}

@article{li2012literature,
  title={Literature survey: domain adaptation algorithms for natural language processing},
  author={Li, Qi},
  journal={Department of Computer Science The Graduate Center, The City University of New York},
  pages={8--10},
  year={2012}
}

@phdthesis{bollmann2018normalization,
  title={Normalization of historical texts with neural network models},
  author={Bollmann, Marcel},
  year={2018},
  school={Ruhr University Bochum, Germany}
}

@article{kestemont2016integrated,
  title={Integrated sequence tagging for medieval Latin using deep representation learning},
  author={Kestemont, Mike and De Gussem, Jeroen},
  journal={arXiv preprint arXiv:1603.01597},
  year={2016}
}

@inproceedings{fang2009latin,
  title={Latin etymologies as features on BNC text categorization},
  author={Fang, Alex Chengyu and Li, Wanyin and Ide, Nancy},
  booktitle={Proceedings of the 23rd Pacific Asia Conference on Language, Information and Computation, Volume 2},
  pages={662--669},
  year={2009}
}

@book{berti2019digital,
  title={Digital Classical Philology: Ancient Greek and Latin in the Digital Revolution},
  author={Berti, Monica},
  volume={10},
  year={2019},
  publisher={Walter de Gruyter GmbH \& Co KG}
}

@article{bamman2020latin,
  title={Latin bert: A contextual language model for classical philology},
  author={Bamman, David and Burns, Patrick J},
  journal={arXiv preprint arXiv:2009.10053},
  year={2020}
}

@inproceedings{turian-etal-2010-word,
    title = "Word Representations: A Simple and General Method for Semi-Supervised Learning",
    author = "Turian, Joseph  and
      Ratinov, Lev-Arie  and
      Bengio, Yoshua",
    booktitle = "Proceedings of the 48th Annual Meeting of the Association for Computational Linguistics",
    month = jul,
    year = "2010",
    address = "Uppsala, Sweden",
    publisher = "Association for Computational Linguistics",
    url = "https://aclanthology.org/P10-1040",
    pages = "384--394",
    }

@book{cltk,
author = {Patrick J. Burns},
editor = {Monica Berti},
doi = {doi:10.1515/9783110599572-010},
url = {https://doi.org/10.1515/9783110599572-010},
title = {Building a Text Analysis Pipeline for Classical Languages},
booktitle = {Digital Classical Philology},
year = {2019},
publisher = {De Gruyter Saur},
pages = {159--176}
}

@inproceedings{Pang2002ThumbsUS,
  title={Thumbs up? Sentiment Classification using Machine Learning Techniques},
  author={Bo Pang and Lillian Lee and Shivakumar Vaithyanathan},
  booktitle={EMNLP},
  year={2002}
}

@inproceedings{w2v,
  author    = {Tom{\'{a}}s Mikolov and
               Kai Chen and
               Greg Corrado and
               Jeffrey Dean},
  editor    = {Yoshua Bengio and
               Yann LeCun},
  title     = {Efficient Estimation of Word Representations in Vector Space},
  booktitle = {1st International Conference on Learning Representations, {ICLR} 2013,
               Scottsdale, Arizona, USA, May 2-4, 2013, Workshop Track Proceedings},
  year      = {2013},
  url       = {http://arxiv.org/abs/1301.3781},
  bibsource = {dblp computer science bibliography, https://dblp.org}
}

@inproceedings{d2v,
author = {Le, Quoc and Mikolov, Tomas},
title = {Distributed Representations of Sentences and Documents},
year = {2014},
publisher = {JMLR.org},
booktitle = {Proceedings of the 31st International Conference on International Conference on Machine Learning - Volume 32},
pages = {II–1188–II–1196},
location = {Beijing, China},
series = {ICML'14}
}

@inproceedings{bert,
    title = "{BERT}: Pre-training of Deep Bidirectional Transformers for Language Understanding",
    author = "Devlin, Jacob  and
      Chang, Ming-Wei  and
      Lee, Kenton  and
      Toutanova, Kristina",
    booktitle = "Proceedings of the 2019 Conference of the North {A}merican Chapter of the Association for Computational Linguistics: Human Language Technologies, Volume 1 (Long and Short Papers)",
    month = jun,
    year = "2019",
    address = "Minneapolis, Minnesota",
    publisher = "Association for Computational Linguistics",
    url = "https://aclanthology.org/N19-1423",
    doi = "10.18653/v1/N19-1423",
    pages = "4171--4186",
}

@MISC{datam,
    author = {Jiawei Han and Micheline Kamber},
    title = {Data Mining: Concepts and Techniques},
    year = {2000}
}

@book{10.5555/1162264,
author = {Bishop, Christopher M.},
title = {Pattern Recognition and Machine Learning (Information Science and Statistics)},
year = {2006},
isbn = {0387310738},
publisher = {Springer-Verlag},
address = {Berlin, Heidelberg}
}

@article{zhang2016semantic,
  title={Semantic NLP-based information extraction from construction regulatory documents for automated compliance checking},
  author={Zhang, Jiansong and El-Gohary, Nora M},
  journal={Journal of Computing in Civil Engineering},
  volume={30},
  number={2},
  pages={04015014},
  year={2016},
  publisher={American Society of Civil Engineers}
}

@article{mcinnes2018umap,
  title={Umap: Uniform manifold approximation and projection for dimension reduction},
  author={McInnes, Leland and Healy, John and Melville, James},
  journal={arXiv preprint arXiv:1802.03426},
  year={2018}
}

@inproceedings{burns2021profiling,
  title={Profiling of Intertextuality in Latin Literature Using Word Embeddings},
  author={Burns, Patrick J and Brofos, James and Li, Kyle and Chaudhuri, Pramit and Dexter, Joseph P},
  booktitle={Proceedings of the 2021 Conference of the North American Chapter of the Association for Computational Linguistics: Human Language Technologies},
  pages={4900--4907},
  year={2021}
}

@article{budkowa1953anonima,
  title={" Anonima tzw. Galla Kronika czyli dzieje ksi{\k{a}}{\.z}{\k{a}}t i w{\l}adc{\'o}w polskich", wyd., wst{\k{e}}pem i komentarzem opatrzy{\l} K. Maleczy{\'n}ski," Pomniki Dziejowe Polski (Monumenta Poloniae Historica) seria II, tom 2:, Krak{\'o}w 1952:[recenzja]},
  author={Budkowa, Zofia and Maleczy{\'n}ski, Karol and Plezia, Marian},
  journal={Przegl{\k{a}}d Historyczny: dwumiesi{\k{e}}cznik naukowy},
  volume={44},
  number={3},
  year={1953}
}

@article{eder2015search,
  title={In search of the author of Chronica Polonorum ascribed to Gallus Anonymus: A stylometric reconnaissance},
  author={Eder, Maciej},
  journal={Acta Poloniae Historica},
  volume={112},
  pages={5--23},
  year={2015}
}
\newpage

\appendix
\vspace{1.5cm}
\section{Images}

\end{document}